%% file: expert_emb.tex
\documentclass{article} 
\usepackage{iclr,times}

\usepackage{amssymb}
\usepackage{amsmath,amsfonts,bm}
\input{symbols.tex}

\usepackage{flushend} 
\usepackage{balance}
 
\usepackage{array,multirow,graphicx}
\usepackage{changes}
\usepackage{booktabs}
\usepackage{graphicx} 
\usepackage{xparse}
\usepackage{hyperref}
\usepackage{url}
\usepackage{subfigure} 
\usepackage{cancel}

\usepackage{sidecap}
\sidecaptionvpos{figure}{t}
\usepackage{verbatimbox}
\usepackage[labelfont=bf]{caption} 
\usepackage{wrapfig}

\usepackage{algorithm}
\usepackage{algpseudocode}
\usepackage{graphics}
\usepackage{epsfig}

\usepackage{color}   
\definecolor{winestain}{rgb}{0.5,0,0}
\definecolor{ocre}{RGB}{51,102,0} 
\definecolor{colorBlue2}{RGB}{200,207,248}
\definecolor{mydarkblue}{rgb}{0,0.08,0.45}
\definecolor{mylightbluetext}{rgb}{0,0.08,0.45}
\usepackage{hyperref}
\hypersetup{
	colorlinks=true, 
	linktoc=all,     
	linkcolor=mydarkblue,  
	anchorcolor=blue,
	citecolor=mydarkblue,
}
\PassOptionsToPackage{numbers, compress}{natbib}
\usepackage[numbers]{natbib}

\renewenvironment{thebibliography}[1]{
	\begin{oldthebibliography}{#1}
		\setlength{\itemsep}{0em}
		\setlength{\parskip}{0.07em}
	}
	{
	\end{oldthebibliography}
}

\makeatletter
\renewcommand\section{\@startsection {section}{1}{\z@}%
	{-0.1\baselineskip}
	{0.1\baselineskip}
	{\normalfont\Large\bfseries}}
\makeatother

\title{Improving embedding with contrastive fine-tuning on small datasets with expert-augmented scores}

\author{
Jun Lu\thanks{Correspondence to: Jun Lu $<$jun.lu.locky@gmail.com$>$.}, \,\,
David Li, \,\,Bill Ding, \,\, Yu Kang\thanks{BA Inc.} 
\\
}

\newtheorem{proof}{Proof}
\newcommand{\BlackBox}{\rule{1.5ex}{1.5ex}}

\begin{document}

\maketitle

\begin{abstract}
This paper presents an approach to improve text embedding models through contrastive fine-tuning on small datasets augmented with expert scores. It focuses on enhancing semantic textual similarity tasks and addressing text retrieval problems. The proposed method uses soft labels derived from expert-augmented scores to fine-tune embedding models, preserving their versatility and ensuring retrieval capability is improved. The paper evaluates the method using a Q\&A dataset from an online shopping website and eight expert models. Results show improved performance over a benchmark model across multiple metrics on various retrieval tasks from the massive text embedding benchmark (MTEB). The method is cost-effective and practical for real-world applications, especially when labeled data is scarce. 

\end{abstract}

\section{Introduction}

Text embedding models are fundamental in natural language processing (NLP), serving as low-dimensional vector representations that capture semantic similarity between texts \citep{aggarwal2012survey, angelov2020top2vec}. They are critical for tasks such as text classification, retrieval, question answering, and dialogue systems. Recent advancements in large language models (LLMs) have spurred interest in retrieval-augmented systems that integrate LLM reasoning with the efficiency of text embeddings.

Two main research directions exist: enhancing performance in semantic textual similarity  (STS) through supervised fine-tuning, normalization, and unsupervised contrastive learning; and addressing text retrieval through dual-encoder architectures and self-supervised pre-training \citep{li2023towards, wang2022text, izacard2020distilling, ren2021rocketqav2, ren2021pair, devlin2018bert, vaswani2017attention}.
Recent studies have aimed to create unified text representation models through large-scale contrastive learning and prompt-based learning \citep{muennighoff2022sgpt}, evaluated on benchmarks like the massive text embedding benchmark (MTEB) \citep{muennighoff2022mteb}, which assesses models across 56 datasets and seven categories.

Innovative models, such as GTE and E5 \citep{li2023towards, wang2022text}, have been developed to address the challenges of creating general-purpose text embeddings. GTE uses multi-stage contrastive learning over diverse datasets, while E5 introduces a progressive learning mechanism to refine embeddings by focusing on zero-shot learning.

Embeddings play a crucial role in retrieval augmented generation (RAG) \citep{lewis2020retrieval, es2023ragas, salemi2024evaluating}. They represent input text and retrieved knowledge sources in a vector space for efficient similarity comparisons. High-quality embeddings improve retrieval accuracy by capturing the semantic and syntactic information of the text. Embeddings can be combined with metadata or context information to enhance retrieval and generation. During the generation process, retrieved knowledge, represented through embeddings, guides the generation of the output text, ensuring it is informed by relevant and accurate information.

In this work, we propose a novel fine-tuning framework for embedding models that utilizes soft labels derived from expert-augmented scores. This approach enables the fine-tuning of embedding models for specific downstream tasks while preserving the versatility of the original embeddings and ensuring that the retrieval capability of the models does not degrade after fine-tuning.
Our framework leverages both expert knowledge and human label information to generate augmented scores that serve as soft labels during the fine-tuning process. These soft labels provide a nuanced guidance signal that helps the model adapt to the specific requirements of the downstream tasks without losing the broad applicability and robustness inherent in the initial embeddings. By doing so, we aim to strike a balance between task-specific performance and the general utility of the embedding models, ensuring they remain effective for retrieval and other tasks even after being fine-tuned.

\section{Related Work}
Text embeddings represent texts using low-dimensional vectors and capture semantic similarity through vector operations. 
There has been a sustained interest in converting texts into compact, low-dimensional representations. Early approaches included methods such as text embedding models like Glove \citep{pennington2014glove} (which lacks context awareness and are thus
commonly labeled as word embedding models), latent semantic analysis (LSA) \citep{deerwester1990indexing} (which achieves embedding  by decomposing a matrix that represents co-occurrences of words and documents to create document embeddings), low-rank matrix factorization with its variants \citep{acharya2019online, lu2024low} (can be incorporated to find the compression of embedding models), and latent Dirichlet allocation (LDA) \citep{blei2003latent} (which uses probabilistic graphical models to infer topic distributions).  

Models like Sentence-BERT (SBERT) \citep{reimers2019sentence} and Sentence-T5 \citep{ni2021sentence} have been fine-tuned on supervised datasets to learn embeddings customized for specific tasks, such as passage retrieval and semantic textual similarity.
On the other hand, methods like inverse cloze task (ICT) \citep{chang2020pre}, random cropping \citep{izacard2021towards}, and neighboring text spans \citep{neelakantan2022text}  have been used to construct text pairs for unsupervised contrastive learning. 

More recently,  GTE (general text embeddings) \citep{li2023towards} is designed to provide robust and versatile text representations for a wide range of NLP and code-related tasks. GTE employs a multi-stage contrastive learning approach, which involves unsupervised pre-training on a large-scale dataset of text pairs sourced from various domains, followed by supervised fine-tuning on high-quality text pairs with human labels from multiple sources.
Despite having a relatively modest parameter count of 110 million, GTE demonstrates exceptional performance, outperforming existing text embedding models and commercial APIs on various benchmarks. 


E5 (EmbEddings from bidirEctional Encoder rEpresentations) \citep{wang2022text} is trained with weak supervision signals from a large-scale curated text pair dataset  CCPairs, and it provides strong off-the-shelf text embeddings that perform exceptionally well in both zero-shot and fine-tuned settings.
The core strength of E5 lies in its ability to transfer effectively across different tasks, making it a versatile choice for applications that require a single-vector representation of texts.
A key feature of E5's training methodology is the use of a contrastive learning framework, which leverages the weak supervision signals from CCPairs to learn robust text representations. This approach allows E5 to capture the nuanced semantic relationships within texts, leading to high-quality embeddings that generalize well to unseen data.

GTE and E5 are examples of multi-stage methods that first train on large-scale datasets and then fine-tune to incorporate human knowledge. 
In contrast, \citet{wang2023improving} introduces a novel and simple method for obtaining high-quality text embeddings using only synthetic data and less than 1,000 training steps. Unlike existing methods, which often involve complex, multi-stage intermediate pretraining with billions of weakly-supervised text pairs, followed by fine-tuning with a few labeled datasets, this method does not require building complex training pipelines or relying on manually collected datasets that are often constrained by task diversity and language coverage. The authors utilize proprietary LLMs to generate diverse synthetic data for hundreds of thousands of text embedding tasks across 93 languages, and then fine-tune open-source decoder-only LLMs on the synthetic data using a standard contrastive loss. Experiments demonstrate that this method achieves strong performance on highly competitive text embedding benchmarks without using any labeled data.

\citet{muennighoff2022mteb} introduces the massive text embedding benchmark (MTEB), which was initially designed to provide a comprehensive evaluation of text embedding models across a diverse set of 8 tasks and 58 datasets covering 112 languages. Since its first publication, the number of datasets included in MTEB has grown to over 500.
MTEB evaluates the performance of text embeddings on a wide range of tasks, including bitext mining, classification, clustering, pair classification, reranking, retrieval, semantic textual similarity, and summarization \citep{conneau2018senteval}.
The authors find that no single text embedding method dominates across all tasks, suggesting that the field has yet to converge on a universal text embedding method that can provide state-of-the-art results on all embedding tasks. MTEB provides open-source code and a public leaderboard to enable easy evaluation of text embedding models and facilitate future research in this area.

\section{Proposed Method}

In this section, we  outline the formulation of contrastive fine-tuning for embedding models using soft labels derived from expert-augmented scores. We begin by discussing the traditional approach of contrastive fine-tuning with hard labels and then introduce our proposed method of contrastive fine-tuning with soft labels.
\subsection{Contrastive Fine-Tuning with Hard Label}

Fine-tuning embedding models using labeled data can incorporate human knowledge into the model, thereby enhancing its performance.
Traditional contrastive fine-tuning methods leverage medium-sized datasets for downstream tasks. For example, in a question-and-answer system (Q\&A system) based on RAG, the same question can be expressed in multiple ways, and the embedding model can be fine-tuned based on these varied formulations of questions (contrastive for irrelevant questions and positive for relevant questions).
In the contrastive fine-tuning framework, contradiction sentences are considered hard negatives and relevant sentences are considered hard positives. 
The loss function for hard labels is the mean squared error (MSE):
$$
\min_\theta \frac{1}{n} \sum_{i=1}^{n}\left(f_\theta(\bq_i)^\top f_\theta(\bp_i) - y_i\right)^2,
$$
where  $f_{\theta}$ represents the fine-tuned model, and $\bq_i$ and $\bp_i$ are the $i$-th query and passages, respectively.
Here, $y_i$ is the hard label; $y_i=1$ if $\bq_i$ and $\bp_i$ are relevant sentences and  $y_i=0$ otherwise.
However, hard labels have their limitations. Although they encapsulate human knowledge, they might introduce overly strict information that can be difficult for embedding models to learn effectively. Furthermore, while fine-tuning on hard labels can benefit specific downstream tasks, it may negatively impact the overall performance of the embedding models, particularly when evaluated on more diverse or open datasets.

\subsection{Contrastive Fine-Tuning with Expert-Augmented Scores}

Suppose we have $K$ expert embedding models, the similarity $s_k$ between a query $\bq$ and a passage $\bp$ for each expert $k$ is defined using the cosine similarity (dot product) of $\bE_{k}(\bq)$ and $\bE_{k}(\bp)$: $s_k = \bE_{k}(\bq)^\top\bE_{k}(\bp)/(\norm{\bE_{k}(\bq)}\cdot \bE_{k}(\bp))$.
The loss function is a soft label based on the $K$ experts:
$$
\min_\theta \frac{1}{n} \sum_{i=1}^{n}\left(f_\theta(\bq_i)^\top f_\theta(\bp_i) - \hat{y}_i\right)^2,
$$
Contrastive fine-tuning  is to distinguish the relevant text pairs from other irrelevant or negative pairs. Therefore, the soft labels can be obtained by 
\begin{equation}\label{equ:soft1}
	\text{Soft-1:}\gap 
	\hat{y}_i=
	\left\{
	\begin{aligned}
		&\max\{s_{1,i}, s_{2,i}, \ldots, s_{K,i} \} &\text{if }& y_i=1\\
		&\min\{s_{1,i}, s_{2,i}, \ldots, s_{K,i} \}             &\text{if }& y_i=0
	\end{aligned}
	\right., 
	\,\,\,  \text{and}\,\,\,  
	s_{k,i}=\frac{\bE_{k}(\bq_i)^\top\bE_{k}(\bp_i)}{\norm{\bE_{k}(\bq_i)}\cdot \norm{\bE_{k}(\bp_i)}}.
\end{equation}
The $K$ experts serve as the source of data distillation, meaning that their collective judgments are distilled into the fine-tuned model. The rationale behind this approach is that hard labels (strict binary relevance) can be too aggressive for the embedding models to achieve. Instead, the expert-augmented scores provide a softer target that is more attainable by the models.

Moreover, since most expert models can differentiate the relevant and irrelevant texts to some extent, the hard label information $\{y_i\}$ can be omitted, and the expert information alone can be utilized such that
\begin{equation}
	\text{Soft-2:}\gap 
	\hat{y}_i= \text{mean} \{s_{1,i}, s_{2,i}, \ldots, s_{K,i} \}.
\end{equation}
That is, the soft labels are regarded as the mean of expert scores. Experiment results show that the Soft-2 scores work slightly worse than Soft-1 models. This is reasonable since we do not use the hard label information.
In cases where the Soft-1 labels are still too aggressive, then we can adopt
\begin{equation}\label{equ:soft3}
	\text{Soft-3:}\gap 
	\hat{y}_i=
	\left\{
	\begin{aligned}
		&\text{mean}\{s_{(1),i}, s_{(2),i}\} &\text{if }& y_i=1\\
		&\text{mean}\{s_{(K-1),i}, s_{(K),i} \}             &\text{if }& y_i=0
	\end{aligned}
	\right., 
\end{equation}
where $\{s_{(1),i}, s_{(2),i}\}$ and $\{s_{(K-1),i}, s_{(K),i} \}$ are the  two highest and two  lowest scores, respectively.
Soft-3 then finds the balance between Soft-1 and Soft-2 labels, incorporating the hard label information but without being too aggressive.

In our experiments, the method requires only a small number of samples for fine-tuning and does not need human labels for the data from multiple sources (see Section~\ref{section:experi}). Overall, the method is resource-efficient and does not require substantial computational resources for fine-tuning.

\section{Experiments}\label{section:experi}
We adopt a Q\&A dataset from an online shopping website. The Q\&A data contains 26 questions with answers provided.
We use open-sourced language models to generate 60 pairs of questions for each question in the Q\&A data. Specifically, we generate 20 similar questions each from  Llama3.1-8B,  Gemma2-9B, and  Qwen2-7B \citep{team2024gemma, yang2024qwen2}~\footnote{The model description can be found, for example, on \url{https://ollama.com/library}.}. 
These are relatively small-sized language models, making the generation of rewritten questions cost-effective.

In other words, for each set of ``question" and ``answer" pair, there are 61 questions and one answer.
Out of these, 21 questions are reserved  for held-out query evaluation, and 40 questions are used to generate pairs of query and passage datasets for fine-tuning.
The answer texts are not used for training and retained  for retrieval evaluation.
Therefore, for each question and answer pair, we extract $40\times 40$ relevant query and passage text pairs (we consider that the order matters). Moreover, within the question and answer pair, `Question1.' and `Question2.' are considered relevant; this indicates that \colorbox{gray!10}{`Question1. And Question2.'} and \colorbox{gray!10}{`Question2.'}, or \colorbox{gray!10}{`Question1. And Question2.'} and \colorbox{gray!10}{`Question1.'} are also relevant. 
This triples the number of relevant text pairs.
For the negative/irrelavant pairs, for each question, we randomly select three questions from a different set of question and answer pair to match the number of positive pairs. As a result, the total number of samples is 249,600 (half positive and half negative), with hard labels provided.
Overall, the effort required to obtain these data was minimal; the source of the dataset is a Q\&A system with 26 questions and answers.

\begin{figure}[t!]
	\centering    
	\vspace{-0.35cm} 
	\subfigtopskip=2pt 
	\subfigbottomskip=2pt 
	\subfigcapskip=-5pt 
	\subfigure[M3e-small.]{\label{fig:exp1}%
		\includegraphics[width=0.33\linewidth]{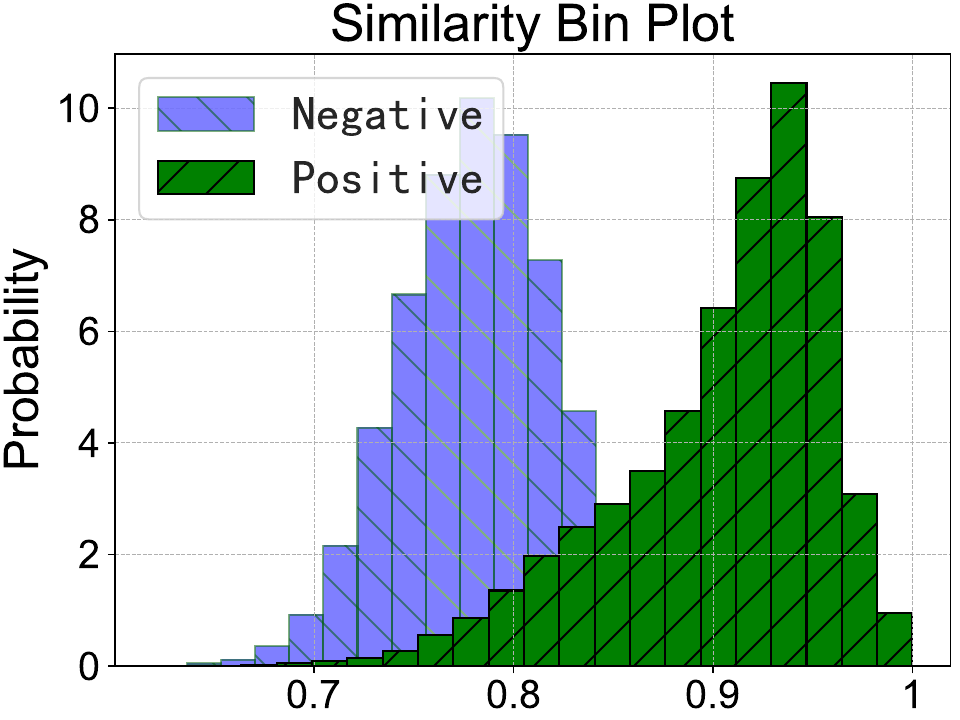}}%
	\subfigure[BCE-embed-base\_v1.]{\label{fig:exp2}%
		\includegraphics[width=0.33\linewidth]{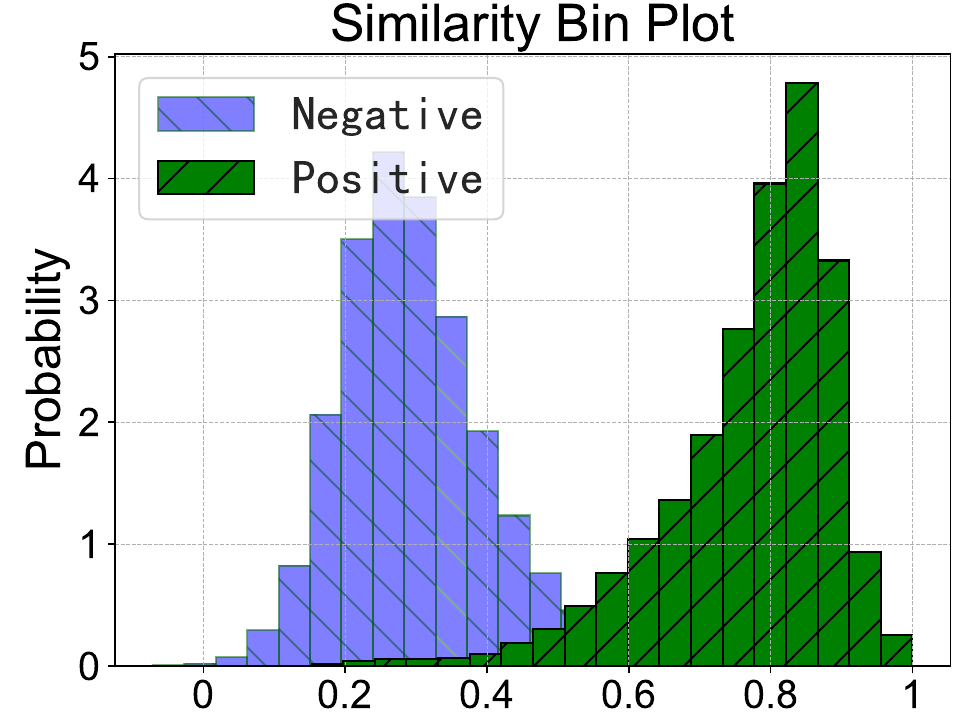}}%
	\subfigure[BGE-large-zh-v1\_5.]{\label{fig:exp3}%
		\includegraphics[width=0.33\linewidth]{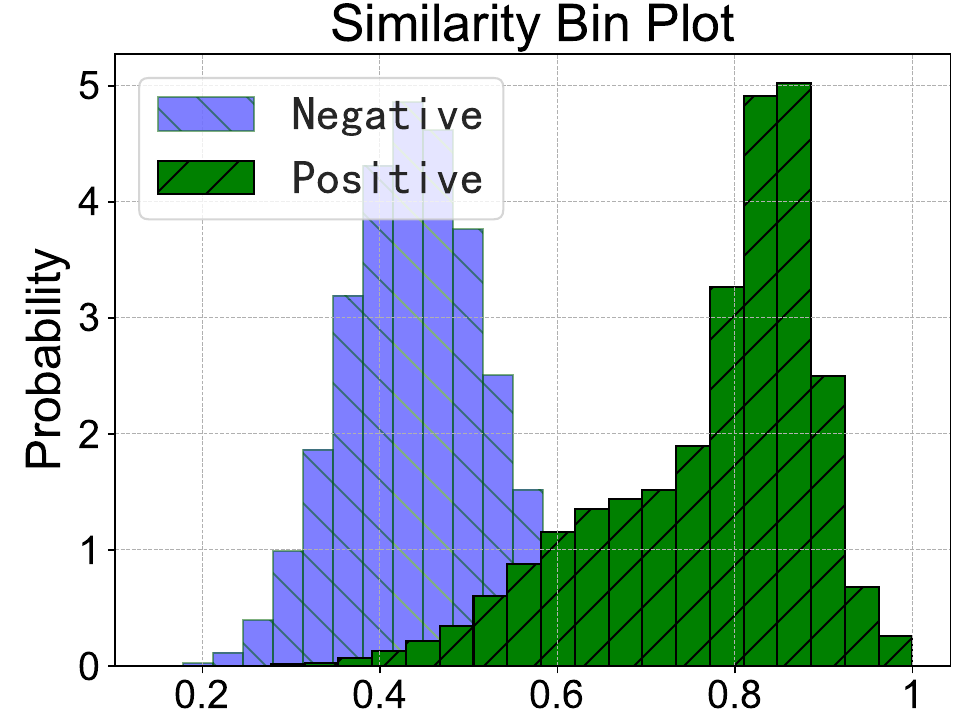}}\\
	\subfigure[Text2vec-base-chinese.]{\label{fig:exp4}%
		\includegraphics[width=0.33\linewidth]{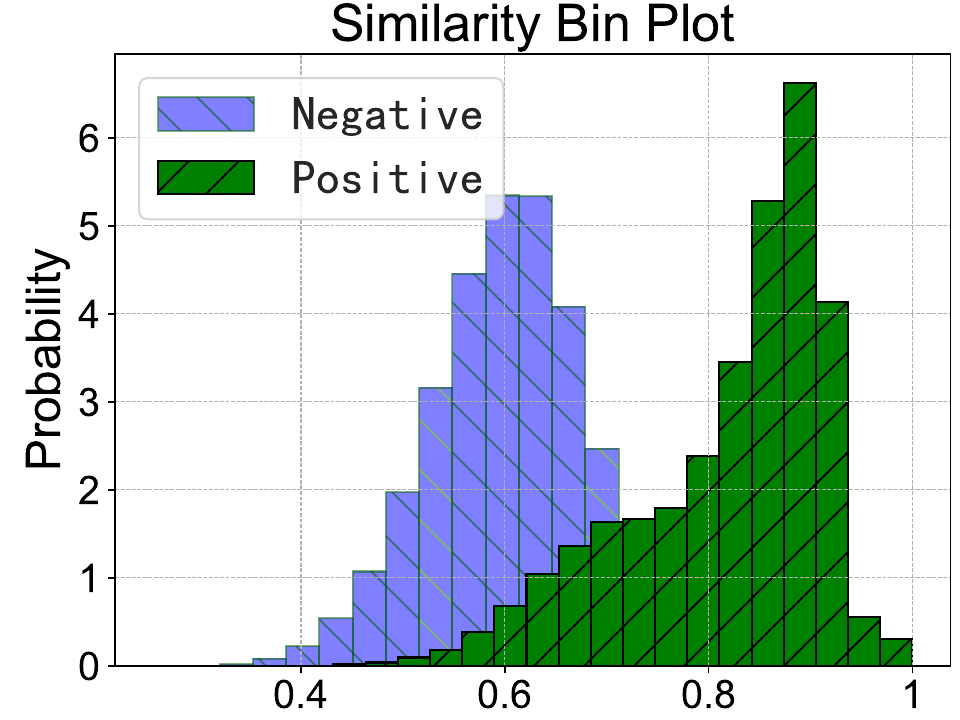}}
	\subfigure[Stella\_en\_1.5B\_v5.]{\label{fig:exp5}%
		\includegraphics[width=0.33\linewidth]{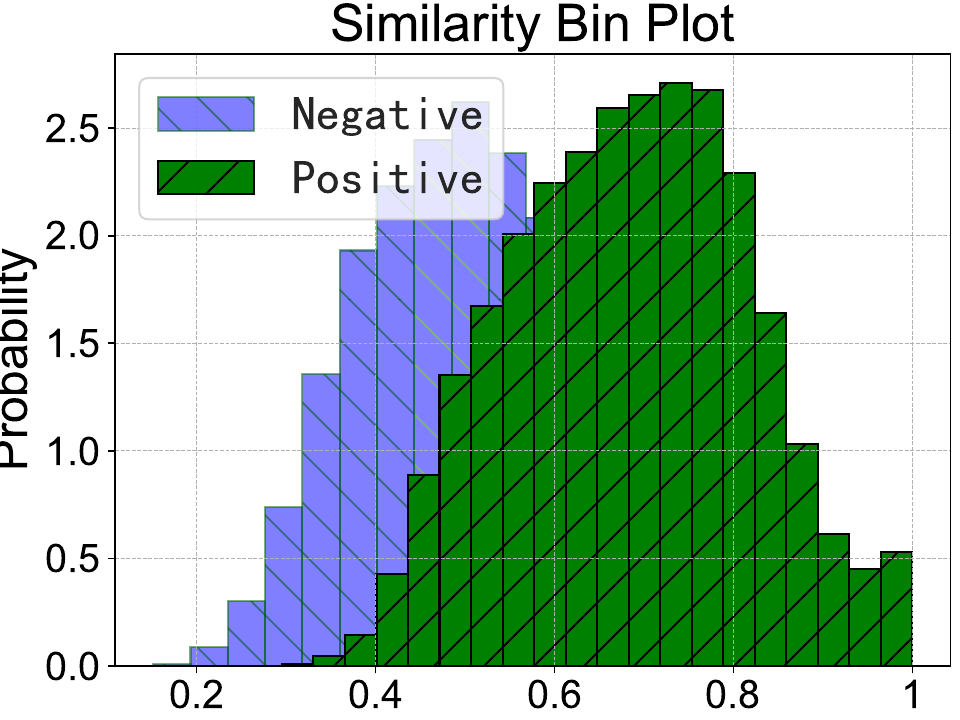}}%
	\subfigure[UAE-Large-V1.]{\label{fig:exp6}%
		\includegraphics[width=0.33\linewidth]{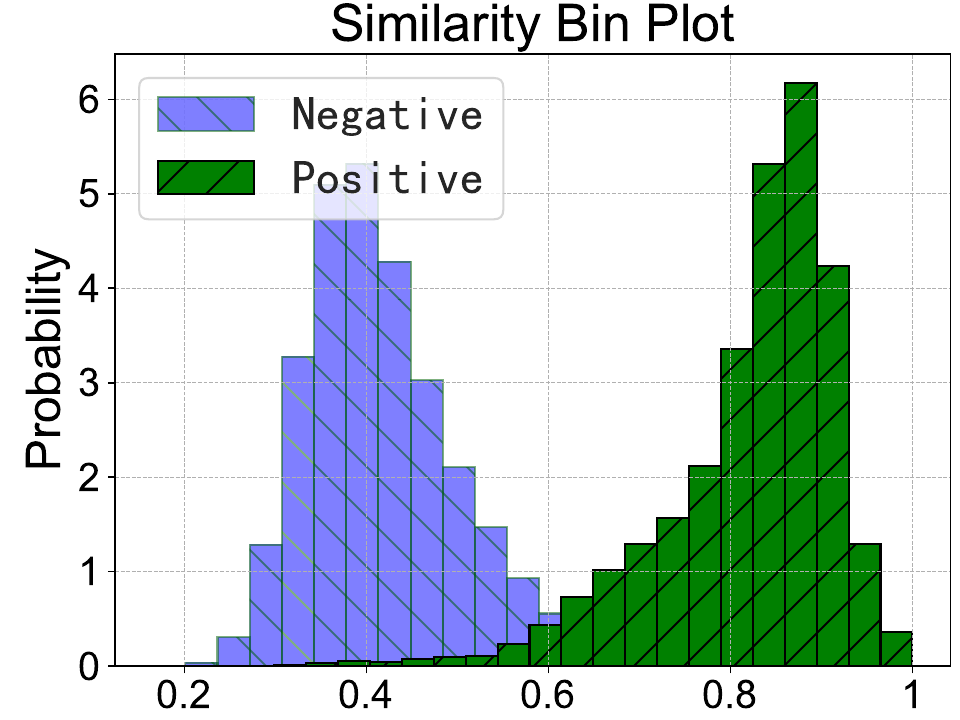}}\\
	\subfigure[Mxbai-embed-large-v1.]{\label{fig:exp7}%
		\includegraphics[width=0.33\linewidth]{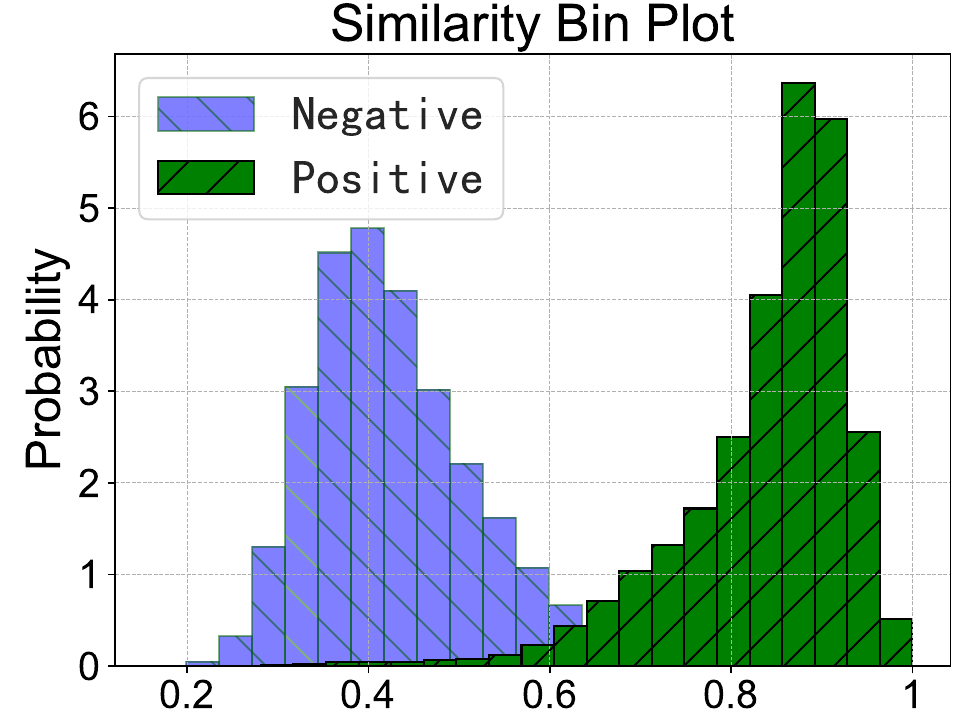}}%
	\subfigure[GTE-large-en-v1.5.]{\label{fig:exp8}%
		\includegraphics[width=0.33\linewidth]{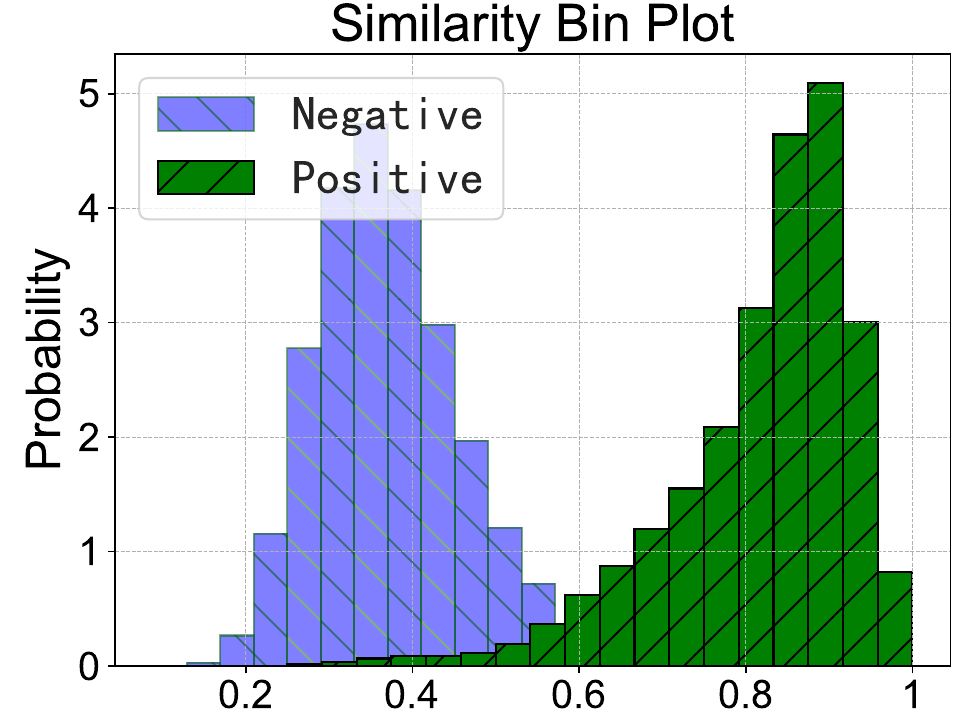}}%
	\subfigure[PR curve.]{\label{fig:expert_pr}%
		\includegraphics[width=0.33\linewidth]{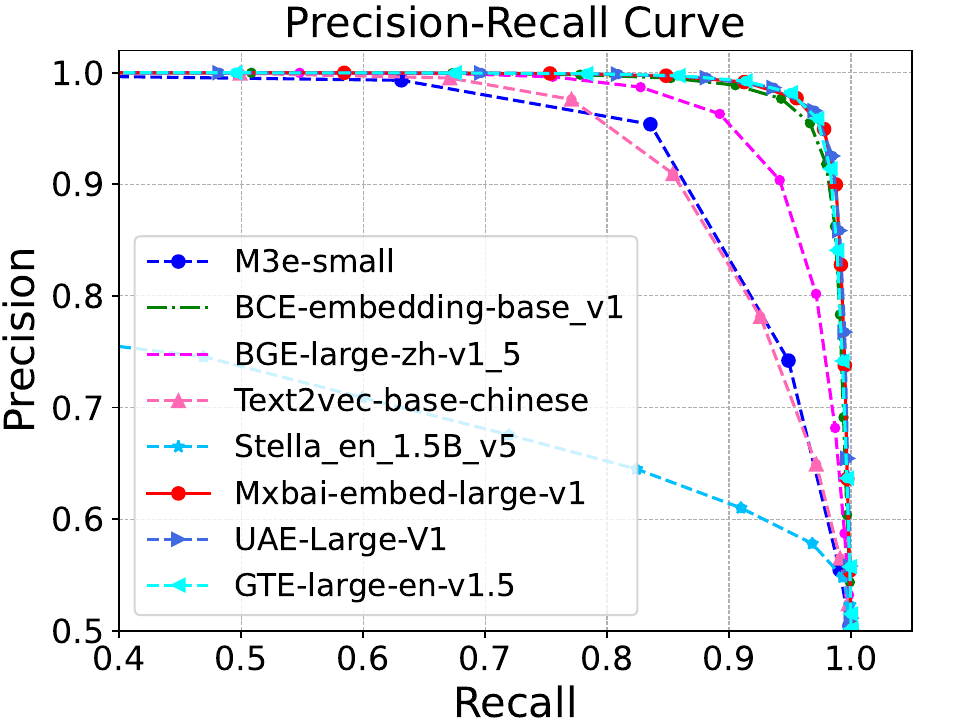}}%
	\caption{Positive and negative label distributions of eight expert models for the given dataset.}
	\label{fig:expt_dists}
\end{figure}

We use eight expert models, namely \textit{M3e-small}, \textit{BCE-embedding-base\_v1}, \textit{BGE-large-zh-v1\_5}
\textit{Text2vec-base-chinese}, \textit{Stella\_en\_1.5B\_v5}, \textit{UAE-Large-V1}, \textit{Mxbai-embed-large-v1}, \textit{GTE-large-en-v1.5}. These models were chosen based on their relatively good performance in the MTEB benchmark and because their sizes are smaller than 2GB~\footnote{\url{https://huggingface.co/spaces/mteb/leaderboard}}. Figure~\ref{fig:exp1} to \ref{fig:exp8} illustrate the positive and negative label distributions for these eight expert models on this Q\&A dataset.
By varying the threshold from 0 to 1, we differentiate the positive and negative samples to obtain both precision and recall values, followed by a precision-recall curve (PR curve) in Figure~\ref{fig:expert_pr}. The PR curve shows that the performance of these models varies.  
The \textit{BCE-embedding-base\_v1}, \textit{Mxbai-embed-large-v1}, \textit{UAE-Large-V1}, and \textit{GTE-large-en-v1.5} models perform relatively better on this dataset;
while \textit{Stella\_en\_1.5B\_v5} performs relatively worse.

For the Soft-1 label defined in Equation~\eqref{equ:soft1}, we define the active set as the set of labels that are equal to the Soft-1 label for each expert model. 
The proportions of the active sets over the total number of samples for 
\textit{M3e-small}, \textit{BCE-embedding-base\_v1}, \textit{BGE-large-zh-v1\_5},
\textit{Text2vec-base-chinese}, \textit{Stella\_en\_1.5B\_v5}, \textit{UAE-Large-V1}, \textit{Mxbai-embed-large-v1}, and \textit{GTE-large-en-v1.5} 
are 40.20\%, 38.23\%,  2.32\%, 1.12\%, 4.79\%, 0.70\%,  2.71\%, and 10.48\%, respectively.
Hence, our approach does not heavily depend on the models that perform well in the PR curve analysis.

The benchmark model selected is \textit{BCE-embedding-base\_v1}, which was recently released  and achieved state-of-the-art results in various datasets \footnote{\url{https://huggingface.co/maidalun1020/bce-embedding-base_v1}. The report on model training has not been released yet.}.
Models are fine-tuned with a batch size of 64 and a learning rate of $3\times 10^{-5}$.
In our experiments, the models show the best performance after 2 epochs, so we only report results for this number of epochs.
In our experiments, there is not a significant difference between the Soft-1 and Soft-3 models, so for brevity, we only present the results of the Soft-1 and Soft-2 models.

\subsection{Evaluation Datasets}
The MTEB Benchmark,  proposed in \citep{muennighoff2022mteb}, is designed for evaluating text embedding models across a wide range of tasks. Although MTEB includes bitext mining datasets that contribute to its various subcategories, the majority of the datasets pertain to retrieval.  
The evaluation metrics used are  mean average precision at 10 (mAP@10), normalized discounted cumulative gain at 10 (nDCG@10), mean reciprocal rank at 10 (mRR@10), and more \citep{muennighoff2022mteb}. 
In this work, we focus on the retrieval subsets.
For retrieval tasks, the default metric suggested by MTEB is nDCG@10, while we show both mAP@10 and  nDCG@10 metrics for evaluation. The mRR@10 shows similar results, and we will not report here for brevity.

Moreover, the dataset we collected contains answer texts, and 21 questions are held out for evaluation for each raw question (with 40 questions used for training). We then use the held-out set and answer texts for retrieval evaluation. 
In this context, each passage contains 40 questions and the corresponding answer. Therefore, for each held-out query (a question out of 21 questions), there is only one relevant passage (see Figure~\ref{fig:inter_intra}). 
The metrics mentioned above are not suitable for this evaluation.
We then plot the similarity of the queries and passages, and vary the thresholds of the similarity scores for evaluating the precisions and recalls.
The metric of the area under precision-recall curve (AUPRC) is considered.

Finally, we use a different dataset from LLaMA-Factory fine-tuning dataset to evaluate the distributional results \footnote{\url{https://github.com/hiyouga/LLaMA-Factory/blob/main/data/alpaca_en_demo.json}.}.
The set contains 1,000 instructions designed for fine-tuning large language models. We evaluate the similarity distributions of texts for different models and compare the symmetric KL divergence of each distribution.

\begin{table*}[t!]
	\scriptsize
	\centering
	\caption{nDCG@10 for various datasets, the larger the better. See \cite{muennighoff2022mteb} for more descriptions on these datasets. The values in brackets are the standard deviations (std). The std of Soft-1 is smaller than that of the Benchmark, indicating greater stability across different tasks. The win rate of Soft-1 over the Benchmark is 50.37\%.}
	\label{tb:mteb_main_ndcg10}
	\begin{tabular}{|c||c|c|c|c|}
		\toprule
		Datasets & Benchmark (BCE) & Soft-1 Label & Soft-2 Label & Hard Label \\\hline 
		\midrule
		AILACasedocs & 16.271 & 24.596 & 24.587 & 21.813   \\ \hline
		AILAStatutes & 25.6 & 20.277 & 19.029 & 26.234   \\ \hline
		ARCChallenge & 8.024 & 7.221 & 7.472 & 7.221   \\ \hline
		AlloprofRetrieval & 27.936 & 30.145 & 30.737 & 28.144   \\ \hline
		AlphaNLI & 3.325 & 13.8 & 15.971 & 3.979   \\ \hline
		AppsRetrieval & 4.813 & 6.231 & 7.256 & 6.001   \\ \hline
		ArguAna & 37.461 & 44.313 & 44.202 & 40.636   \\ \hline
		BSARDRetrieval & 4.236 & 9.699 & 10.161 & 5.793   \\ \hline
		BelebeleRetrieval & 60.589 & 60.928 & 56.978 & 57.214   \\ \hline
		CodeEditSearchRetrieval & 40.907 & 43.266 & 41.876 & 36.574   \\ \hline
		CodeFeedbackMT & 27.305 & 35.992 & 35.987 & 33.398   \\ \hline
		CodeFeedbackST & 55.711 & 69.684 & 67.556 & 59.338   \\ \hline
		CodeSearchNetCCRetrieval & 47.772 & 59.999 & 58.784 & 35.701   \\ \hline
		CodeSearchNetRetrieval & 88.494 & 80.598 & 80.034 & 81.016   \\ \hline
		CodeTransOceanContest & 55.59 & 59.969 & 60.571 & 55.538   \\ \hline
		CodeTransOceanDL & 32.557 & 24.528 & 24.842 & 33.991   \\ \hline
		CosQA & 19.373 & 21.382 & 16.685 & 13.64   \\ \hline
		CrossLingualSemanticDiscriminationWMT19 & 88.858 & 93.409 & 93.3 & 80.44   \\ \hline
		CrossLingualSemanticDiscriminationWMT21 & 89.615 & 93.506 & 93.082 & 83.7   \\ \hline
		DanFeverRetrieval & 33.357 & 34.012 & 34.279 & 33.72   \\ \hline
		EstQA & 70.694 & 65.822 & 63.985 & 65.377   \\ \hline
		FQuADRetrieval & 69.406 & 66.137 & 65.68 & 58.099   \\ \hline
		GeorgianFAQRetrieval & 50.66 & 47.539 & 44.789 & 45.426   \\ \hline
		GerDaLIR & 3.681 & 4.938 & 4.982 & 4.632   \\ \hline
		GerDaLIRSmall & 8.751 & 11.225 & 11.894 & 11.709   \\ \hline
		GermanDPR & 71.17 & 66.826 & 68.112 & 69.006   \\ \hline
		GermanGovServiceRetrieval & 81.043 & 81.004 & 78.707 & 75.581   \\ \hline
		GermanQuAD-Retrieval & 87.875 & 84.649 & 83.224 & 86.49   \\ \hline
		GreekCivicsQA & 81.502 & 80.508 & 79.53 & 73.83   \\ \hline
		HunSum2AbstractiveRetrieval & 84.393 & 82.987 & 82.77 & 82.674   \\ \hline
		IndicQARetrieval & 44.098 & 41.736 & 39.551 & 39.65   \\ \hline
		JaGovFaqsRetrieval & 55.853 & 57.679 & 54.304 & 53.184   \\ \hline
		JaQuADRetrieval & 47.412 & 51.135 & 49.832 & 45.559   \\ \hline
		Ko-StrategyQA & 58.117 & 52.942 & 50.403 & 55.882   \\ \hline
		LegalQuAD & 26.595 & 27.932 & 29.789 & 29.439   \\ \hline
		MLQARetrieval & 41.139 & 36.237 & 33.176 & 33.539   \\ \hline
		MintakaRetrieval & 10.222 & 8.162 & 11.654 & 10.559   \\ \hline
		MultiLongDocRetrieval & 13.621 & 12.322 & 11.003 & 14.177   \\ \hline
		NLPJournalAbsIntroRetrieval & 82.299 & 83.245 & 83.943 & 75.827   \\ \hline
		NLPJournalTitleAbsRetrieval & 91.733 & 88.784 & 90.754 & 76.062   \\ \hline
		NLPJournalTitleIntroRetrieval & 69.344 & 65.921 & 68.423 & 54.441   \\ \hline
		Quora-PL & 74.362 & 73.308 & 71.973 & 68.605   \\ \hline
		RARbCode & 48.251 & 36.934 & 31.102 & 39.684   \\ \hline
		RARbMath & 8.739 & 47.508 & 49.866 & 26.315   \\ \hline
		SCIDOCS & 11.395 & 10.601 & 11.177 & 11.51   \\ \hline
		SIQA & 0.171 & 1.886 & 1.745 & 0.543   \\ \hline
		SciFact & 56.961 & 52.465 & 53.703 & 53.034   \\ \hline
		SpartQA & 3.39 & 5.431 & 9.345 & 2.585   \\ \hline
		StackOverflowQA & 75.758 & 64.69 & 65.365 & 68.953   \\ \hline
		SyntecRetrieval & 66.987 & 72.362 & 72.342 & 64.349   \\ \hline
		SyntheticText2SQL & 48.335 & 45.494 & 45.23 & 35.547   \\ \hline
		TRECCOVID & 50.099 & 50.257 & 47.588 & 44.389   \\ \hline
		TRECCOVID-PL & 40.363 & 37.302 & 34.739 & 37.292   \\ \hline
		TV2Nordretrieval & 79.676 & 79.661 & 77.591 & 77.287   \\ \hline
		TempReasonL1 & 1.025 & 0.78 & 0.924 & 0.863   \\ \hline
		TempReasonL2Context & 1.019 & 3.386 & 3.433 & 2.923   \\ \hline
		TempReasonL2Fact & 3.667 & 6.115 & 6.489 & 6.006   \\ \hline
		TempReasonL2Pure & 0.548 & 0.292 & 0.234 & 0.414   \\ \hline
		TempReasonL3Context & 1.596 & 4.184 & 4.054 & 3.569   \\ \hline
		TempReasonL3Fact & 5.942 & 7.36 & 7.863 & 7.183   \\ \hline
		TempReasonL3Pure & 4.573 & 4.946 & 4.026 & 4.642   \\ \hline
		TopiOCQA & 9.061 & 8.957 & 9.228 & 9.775   \\ \hline
		Touche2020 & 16.518 & 13.668 & 12.927 & 14.448   \\ \hline
		TwitterHjerneRetrieval & 53.037 & 43.303 & 45.405 & 56.55   \\ \hline
		WinoGrande & 0.0 & 18.968 & 25.445 & 0.626   \\ \hline
		\hline 
		Average & 39.675 (29.963) & \textbf{40.633} (28.552) & 40.334 (28.167) & 37.574 (\textbf{27.081})   \\  
		\hline 
		\bottomrule
	\end{tabular}
\end{table*}

\begin{table}[htp]
	\scriptsize
	\centering
	\caption{mAP@10 for various datasets, the larger the better. See \cite{muennighoff2022mteb} for more descriptions on these datasets. The values in brackets are the standard deviations (std).  The std of Soft-1 is smaller than that of the Benchmark, indicating greater stability across different tasks. The win rate of Soft-1 over the Benchmark is 55.38\%.}
	\label{tb:mteb_main_map10}
	\begin{tabular}{|c||c|c|c|c|}
		\toprule
		Datasets & Benchmark (BCE) & Soft-1 Label & Soft-2 Label & Hard Label \\\hline 
		\midrule
		AILACasedocs & 9.21 & 16.445 & 17.97 & 13.749   \\ \hline
		AILAStatutes & 15.117 & 11.844 & 9.773 & 15.607   \\ \hline
		ARCChallenge & 6.184 & 5.477 & 5.654 & 5.393   \\ \hline
		AlloprofRetrieval & 23.256 & 24.758 & 25.686 & 23.123   \\ \hline
		AlphaNLI & 2.67 & 11.646 & 13.726 & 3.084   \\ \hline
		AppsRetrieval & 4.017 & 5.184 & 6.095 & 5.067   \\ \hline
		ArguAna & 30.588 & 35.959 & 35.892 & 33.19   \\ \hline
		BSARDRetrieval & 3.049 & 7.445 & 7.758 & 4.425   \\ \hline
		BelebeleRetrieval & 55.729 & 56.366 & 51.915 & 52.509   \\ \hline
		CodeEditSearchRetrieval & 37.412 & 39.591 & 38.143 & 33.195   \\ \hline
		CodeFeedbackMT & 23.867 & 32.652 & 32.703 & 29.709   \\ \hline
		CodeFeedbackST & 50.525 & 64.973 & 62.731 & 54.205   \\ \hline
		CodeSearchNetCCRetrieval & 43.033 & 55.399 & 54.369 & 31.743   \\ \hline
		CodeSearchNetRetrieval & 85.781 & 76.519 & 76.022 & 77.056   \\ \hline
		CodeTransOceanContest & 50.976 & 56.021 & 56.451 & 51.7   \\ \hline
		CodeTransOceanDL & 19.937 & 15.744 & 16.098 & 21.395   \\ \hline
		CosQA & 15.49 & 16.562 & 12.897 & 10.435   \\ \hline
		CrossLingualSemanticDiscriminationWMT19 & 85.735 & 91.232 & 91.073 & 76.096   \\ \hline
		CrossLingualSemanticDiscriminationWMT21 & 86.317 & 91.316 & 90.759 & 79.073   \\ \hline
		DanFeverRetrieval & 30.705 & 31.516 & 31.739 & 31.042   \\ \hline
		EstQA & 64.272 & 59.333 & 57.828 & 58.465   \\ \hline
		FQuADRetrieval & 64.136 & 60.788 & 60.218 & 52.019   \\ \hline
		GeorgianFAQRetrieval & 45.927 & 42.812 & 39.888 & 40.409   \\ \hline
		GerDaLIR & 2.984 & 4.038 & 3.994 & 3.702   \\ \hline
		GerDaLIRSmall & 7.141 & 9.266 & 9.801 & 9.574   \\ \hline
		GermanDPR & 64.46 & 59.705 & 61.485 & 62.163   \\ \hline
		GermanGovServiceRetrieval & 75.437 & 75.547 & 72.481 & 68.443   \\ \hline
		GermanQuAD-Retrieval & 85.367 & 81.504 & 79.792 & 83.637   \\ \hline
		GreekCivicsQA & 77.457 & 76.195 & 75.229 & 68.726   \\ \hline
		HunSum2AbstractiveRetrieval & 82.114 & 80.405 & 80.108 & 80.028   \\ \hline
		IndicQARetrieval & 38.454 & 36.58 & 34.405 & 34.29   \\ \hline
		JaGovFaqsRetrieval & 51.284 & 52.884 & 49.548 & 48.58   \\ \hline
		JaQuADRetrieval & 37.276 & 40.362 & 39.321 & 35.311   \\ \hline
		Ko-StrategyQA & 51.167 & 45.674 & 42.931 & 48.945   \\ \hline
		LegalQuAD & 22.18 & 23.868 & 24.999 & 24.881   \\ \hline
		MLQARetrieval & 37.099 & 32.345 & 29.416 & 29.773   \\ \hline
		MintakaRetrieval & 8.429 & 6.733 & 9.743 & 8.674   \\ \hline
		MultiLongDocRetrieval & 11.681 & 10.012 & 9.146 & 12.286   \\ \hline
		NLPJournalAbsIntroRetrieval & 79.125 & 80.517 & 80.84 & 72.374   \\ \hline
		NLPJournalTitleAbsRetrieval & 89.589 & 85.956 & 88.514 & 71.591   \\ \hline
		NLPJournalTitleIntroRetrieval & 64.653 & 61.348 & 63.581 & 49.824   \\ \hline
		Quora-PL & 69.412 & 68.232 & 66.971 & 63.78   \\ \hline
		RARbCode & 42.661 & 32.289 & 26.48 & 34.827   \\ \hline
		RARbMath & 7.358 & 44.916 & 47.206 & 23.856   \\ \hline
		SCIDOCS & 6.264 & 5.762 & 6.16 & 6.278   \\ \hline
		SIQA & 0.097 & 1.457 & 1.369 & 0.427   \\ \hline
		SciFact & 52.262 & 48.234 & 48.73 & 47.923   \\ \hline
		SpartQA & 1.73 & 2.847 & 5.576 & 1.18   \\ \hline
		StackOverflowQA & 72.487 & 61.65 & 62.558 & 65.687   \\ \hline
		SyntecRetrieval & 61.604 & 65.865 & 65.705 & 57.292   \\ \hline
		SyntheticText2SQL & 37.538 & 35.359 & 35.064 & 26.858   \\ \hline
		TRECCOVID & 1.041 & 1.096 & 0.932 & 1.0   \\ \hline
		TRECCOVID-PL & 0.845 & 0.735 & 0.698 & 0.732   \\ \hline
		TV2Nordretrieval & 76.553 & 76.558 & 74.29 & 73.78   \\ \hline
		TempReasonL1 & 0.645 & 0.504 & 0.621 & 0.517   \\ \hline
		TempReasonL2Context & 0.67 & 2.431 & 2.457 & 2.001   \\ \hline
		TempReasonL2Fact & 2.604 & 4.5 & 4.78 & 4.408   \\ \hline
		TempReasonL2Pure & 0.383 & 0.212 & 0.171 & 0.284   \\ \hline
		TempReasonL3Context & 1.038 & 2.926 & 2.896 & 2.523   \\ \hline
		TempReasonL3Fact & 4.089 & 5.316 & 5.756 & 5.148   \\ \hline
		TempReasonL3Pure & 2.797 & 3.02 & 2.444 & 2.89   \\ \hline
		TopiOCQA & 7.376 & 7.381 & 7.863 & 7.728   \\ \hline
		Touche2020 & 5.811 & 4.47 & 3.956 & 5.025   \\ \hline
		TwitterHjerneRetrieval & 42.157 & 33.299 & 35.003 & 45.808   \\ \hline
		WinoGrande & 0.0 & 14.417 & 19.208 & 0.359   \\ \hline
		\hline 
		Average & 34.419 (29.693) & \textbf{35.323} (28.587) & 35.04 (28.221) & 32.243 (\textbf{26.585})   \\  
		\hline 
		\bottomrule
	\end{tabular}
\end{table}

\subsection{Results on MTEB}
Table~\ref{tb:mteb_main_ndcg10} and~\ref{tb:mteb_main_map10} present  the evaluation of nDCG@10 and mAP@10 metrics, respectively, for different models across various datasets from MTEB retrieval tasks. 
The average nDCG@10 scores for Benchmark, Soft-1, Soft-2, and Hard label models are 39.675, \textbf{40.633}, 40.334, and 37.574, respectively, with  standard deviations of 29.963, 28.552, 28.167, and 27.081, respectively. 
And the average mAP@10 for Benchmark, Soft-1, Soft-2, and Hard label models are 34.419, \textbf{35.323}, 35.04, and 32.243, respectively, with standard deviations of 29.693, 28.587, 28.221, and 26.585, respectively.
The win rate of Soft-1 over the Benchmark is 50.37\% in terms of nDCG@10, and is 55.38\% with respect to mAP@10.
This again confirms that no single text embedding method dominates across all tasks \citep{muennighoff2022mteb}.
The Soft-1 and Soft-2 models demonstrate promising results with higher scores and smaller standard deviations compared to the Benchmark model, suggesting they perform well across various datasets and their performance is consistently stable.
The Hard label model, on the other hand, has worse nDCG@10 and mAP@10 scores compared to the Benchmark; although it has a smaller standard deviation.

The improvement seen in the fine-tuning with Soft-1 and Soft-2 labels might be attributed to the reduced anisotropy in the fine-tuned models (meaning the text embeddings occupy a larger cone in the vector space after fine-tuning). 
This property is further supported by the results on the held-out set: the Soft-1 and Soft-2 models have better results in terms of area under precision-recall (PR) curve (see Section~\ref{section:heldouts}). The text embeddings of irrelevant pairs are then distributed across  a wider range of the vector space.

\begin{figure}[t!]
	\centering    
	\vspace{-0.35cm} 
	\subfigtopskip=2pt 
	\subfigbottomskip=2pt 
	\subfigcapskip=-5pt 
	\subfigure[Benchmark.]{\label{fig:expert_prcurive_bin_benchmark}%
		\includegraphics[width=0.25\linewidth]{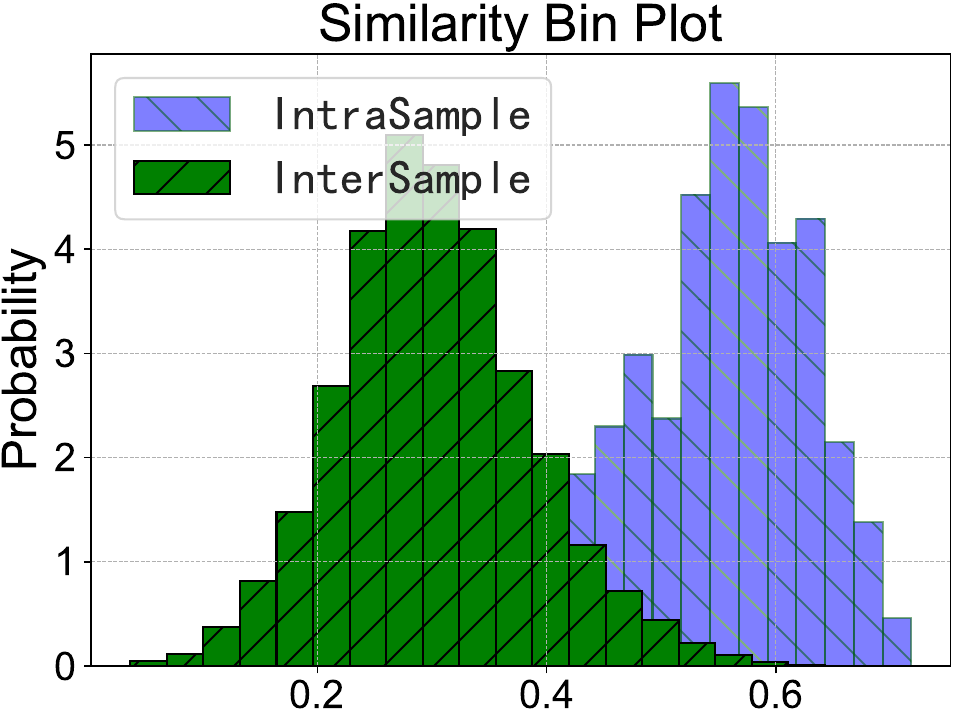}}%
	\subfigure[Hard.]{\label{fig:expert_prcurive_bin_hard}%
		\includegraphics[width=0.25\linewidth]{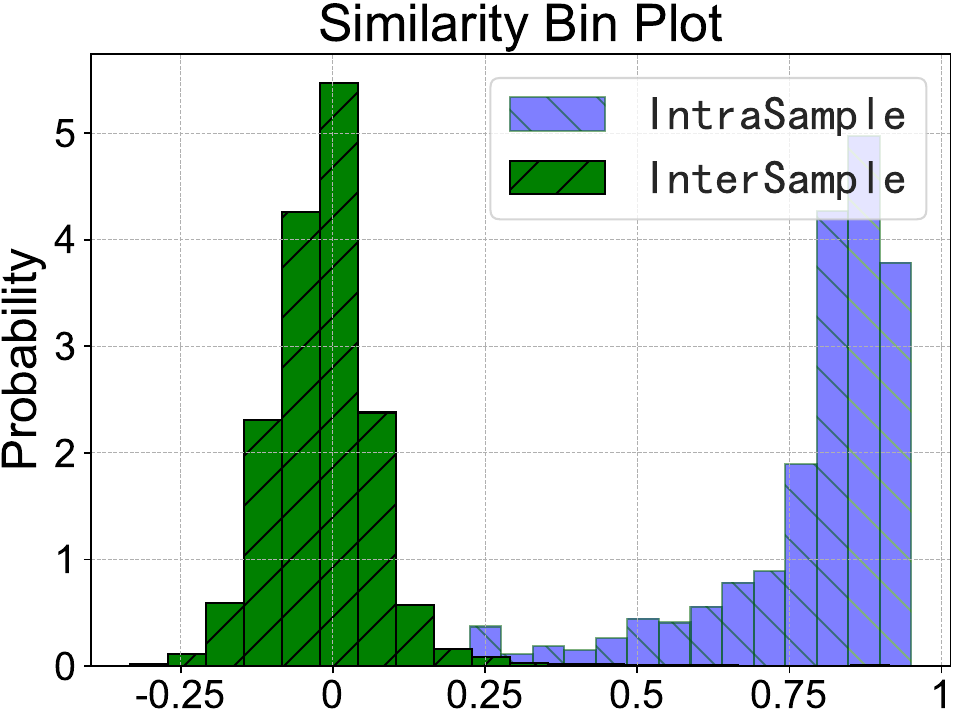}}%
	\subfigure[Soft-1.]{\label{fig:expert_prcurive_bin_soft1}%
		\includegraphics[width=0.25\linewidth]{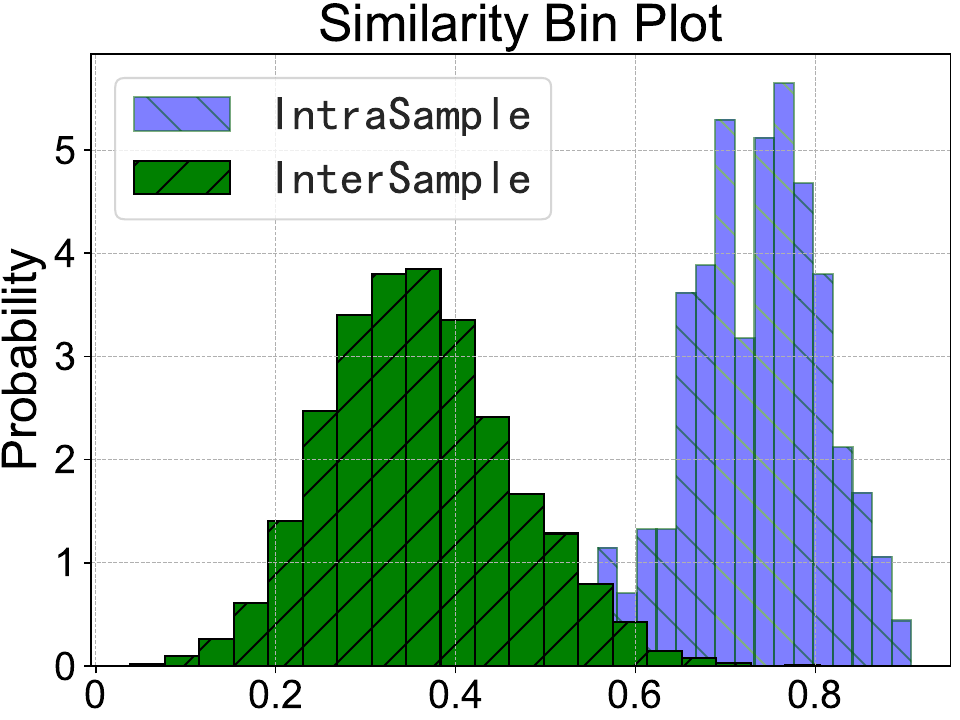}}%
	\subfigure[Soft-2.]{\label{fig:expert_prcurive_bin_soft2}%
		\includegraphics[width=0.25\linewidth]{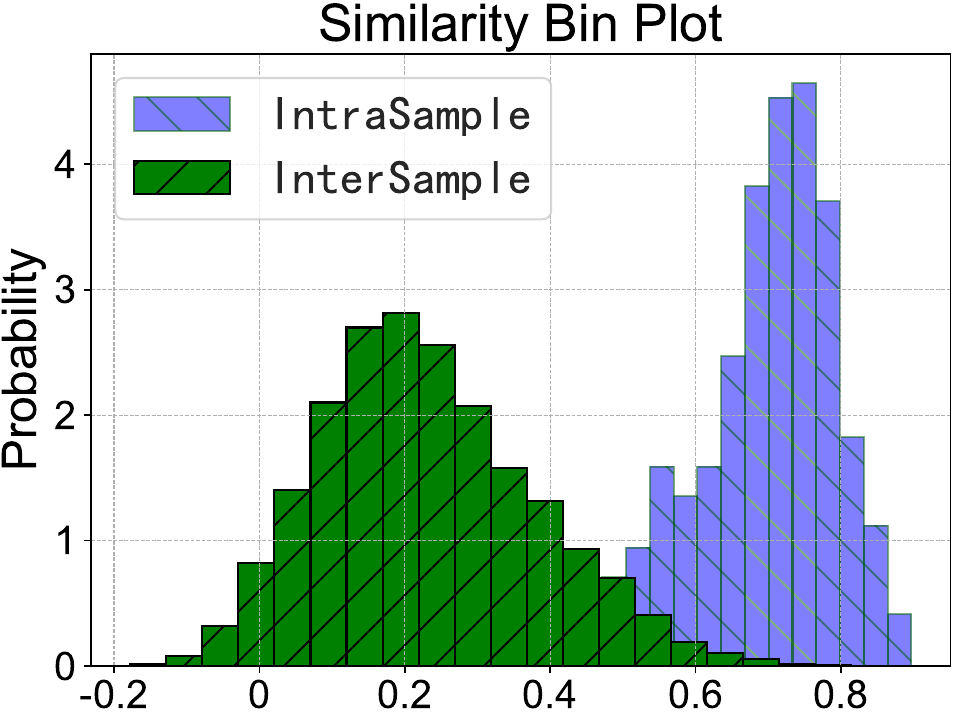}}%
	\caption{Distributions of the IntraSample and InterSample for Benchmark, Hard label, Soft-1, and Soft-2, respectively.
		All models exhibit a distinct separation between IntraSample and InterSample distributions. In this scenario, the Hard label model appears to perform the best due to a more pronounced difference between the two modalities, while in terms of AUPRC, the Soft-1 performs best (Table~\ref{tb:exp_auprc}).}
	\label{fig:expert_prcurive_bin}
\end{figure}

\subsection{Results on Held-Out Set}\label{section:heldouts}
\begin{figure}[t!]
	\begin{center}
		\begin{minipage}[t]{0.8\linewidth}
			\centering
			\vspace{-0.35cm} 
			\subfigtopskip=2pt 
			\subfigbottomskip=2pt 
			\subfigcapskip=-5pt 
			{\includegraphics[width=0.9\textwidth]{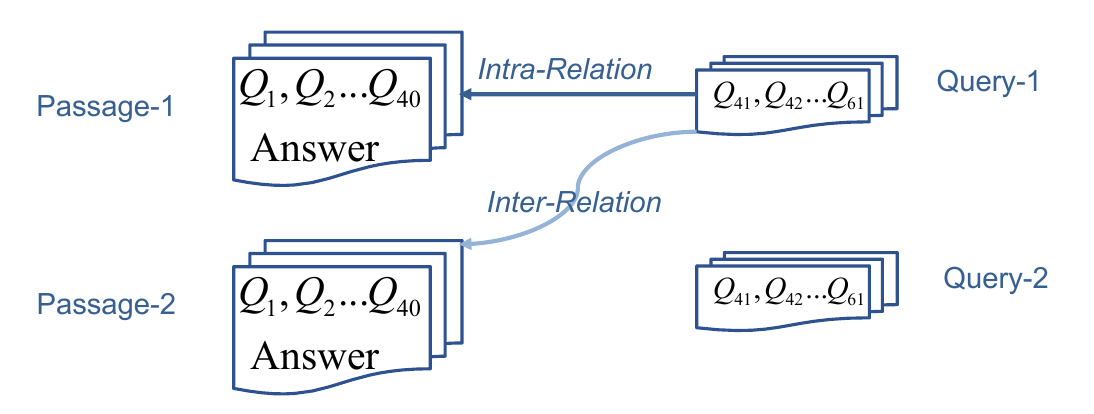}} \\ \vspace{-0.10cm}
		\end{minipage}
		\centering
		\caption{Diagram illustrating the inter- and intra-relationship between different queries and passages.}
		\label{fig:inter_intra}
	\end{center}
	\vspace{-.2in}
\end{figure}

Building upon the  held-out Q\&A dataset, since for each held-out question, there is only one related passage, we examine the similarity between the embedding vectors for inter- and intra-questions; see Figure~\ref{fig:inter_intra}. 
The goal of a retrieval process for a Q\&A system is to determine whether the embeddings can help differentiate related passages from irrelevant ones, so that the system can use the related passage for downstream tasks (such as RAG).
Define further the term \textit{``IntraSample"} as the similarity between a question from the $i$-th query (there are 21 questions for each query) and the $i$-th passage, and \textit{``InterSample"} as the similarity between a question from the $i$-th query and the $j$-th passage ($j\neq i$, and since there are 26 questions in the Q\&A dataset, $i,j\in\{1,2,\ldots,26\}$). 
The method is frequently used in recommender systems; see, for example, \cite{lu2021numerical}. Figure~\ref{fig:expert_prcurive_bin_benchmark}, \ref{fig:expert_prcurive_bin_hard}, \ref{fig:expert_prcurive_bin_soft1}, and \ref{fig:expert_prcurive_bin_soft2} depict the bin plots of the distributions of IntraSample and InterSample under cosine similarity for the Benchmark, Hard label, Soft-1, and Soft-2 models, respectively.
\noindent
\begin{center}
	\begin{minipage}{0.85\textwidth}
		\begin{minipage}[b]{0.56\textwidth}
			\centering
			\includegraphics[width=1\textwidth]{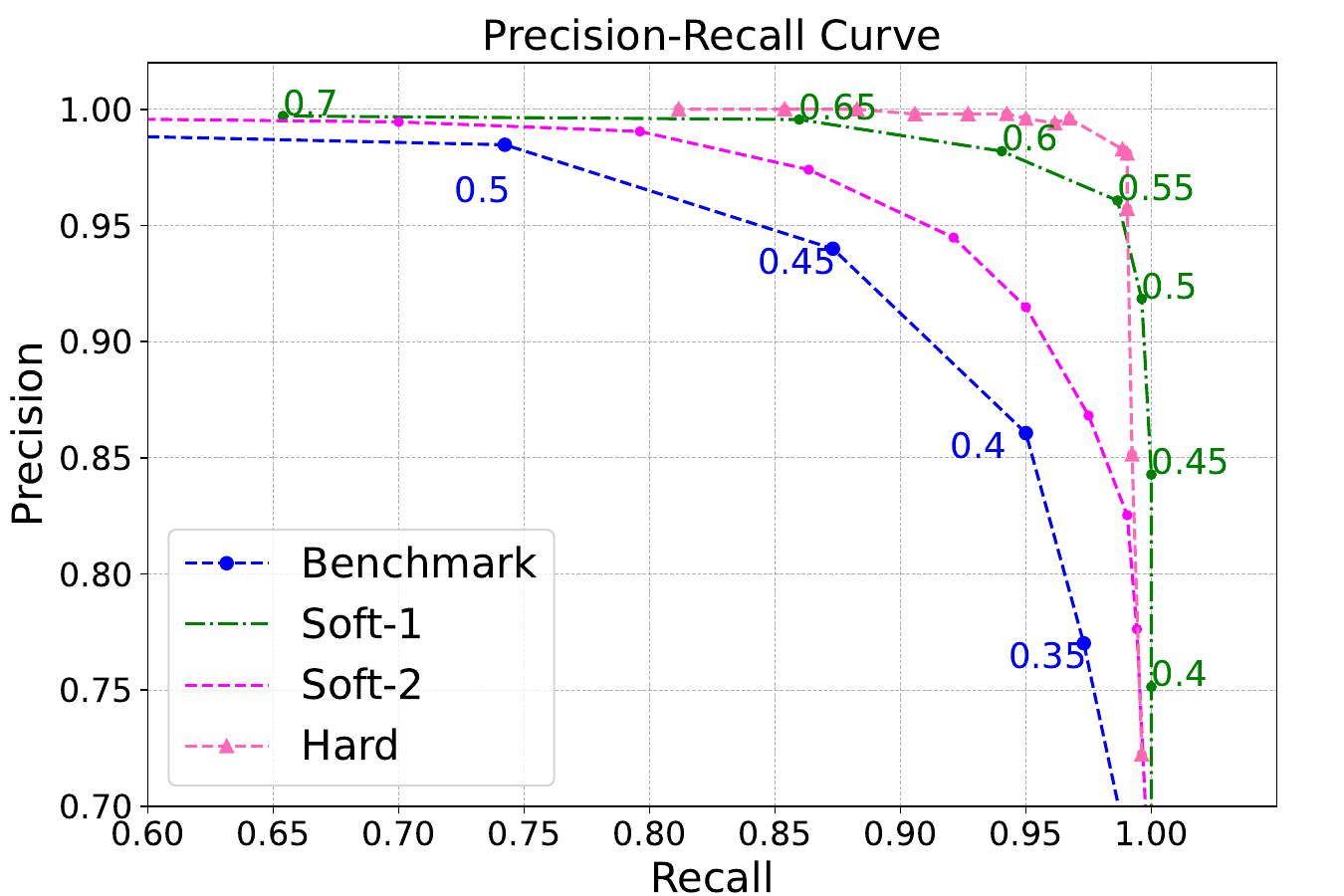}
			\captionof{figure}{PR curves of held-out set analysis for various methods. Threshold values (see Figure~\ref{fig:expert_prcurive_bin}) are only shown for Benchmark and Soft-1 for conciseness. }
			\label{fig:rag_prcurve}
		\end{minipage}
		\hfill
		\begin{minipage}[b]{0.4\textwidth}
			\centering
			\begin{tabular}{|c|c|}
				\toprule
				Models & AUPRC  \\
				\midrule
				Benchmark (BCE) & 0.9708   \\ 
				Soft-1 Label & \textbf{0.9960}   \\ 
				Soft-2 Label & 0.9855   \\ 
				Hard Label & 0.9947   \\ \hline
				\bottomrule
			\end{tabular}
			\captionof{table}{Area under the PR curve (AUPRC, the larger the better, with a maximum value of 1) of the held-out set analysis for various methods. The proposed Soft-1 model obtains a better result compared to the Benchmark and Hard label models, with a value of 0.9960.}
			\label{tb:exp_auprc}
		\end{minipage}
	\end{minipage}
\end{center}
\noindent
\noindent
In all scenarios, a clear distinction is observed between the distributions of the IntraSample and InterSample data, showing that general embedding models  can effectively find the related passages to some extent.
Figure~\ref{fig:rag_prcurve} shows the \textit{precision-recall (PR) curve} for these scenarios (by varying the thresholds to find the precisions and recalls of InterSample under different thresholds), and Table~\ref{tb:exp_auprc} displays the area under the PR curve (AUPRC, the larger the better with a maximum value of 1), where we find the Soft-1 model performs best in terms of AUPRC with a value of 0.9960.
The Hard label model achieves the best precision for various recall values, which is reasonable given that the Hard label model fine-tunes the label to a significant extent, sacrificing its versatility in open datasets (as shown in the results of MTEB benchmark). Moreover, there is a substantial  drop in the precision value when decreasing the threshold from 0.4 to 0.1  in the Hard label model, resulting a smaller AUPRC  compared to the Soft-1 model.
This suggests that the overlap between the IntraSample and InterSample is more pronounced in the Hard label model than in the Soft-1 model, making it more challenging to differentiate the samples within this overlap.

\begin{center}
\begin{minipage}{1\textwidth}
	\begin{minipage}[b]{0.5\textwidth}
		\centering
		\includegraphics[width=1\textwidth]{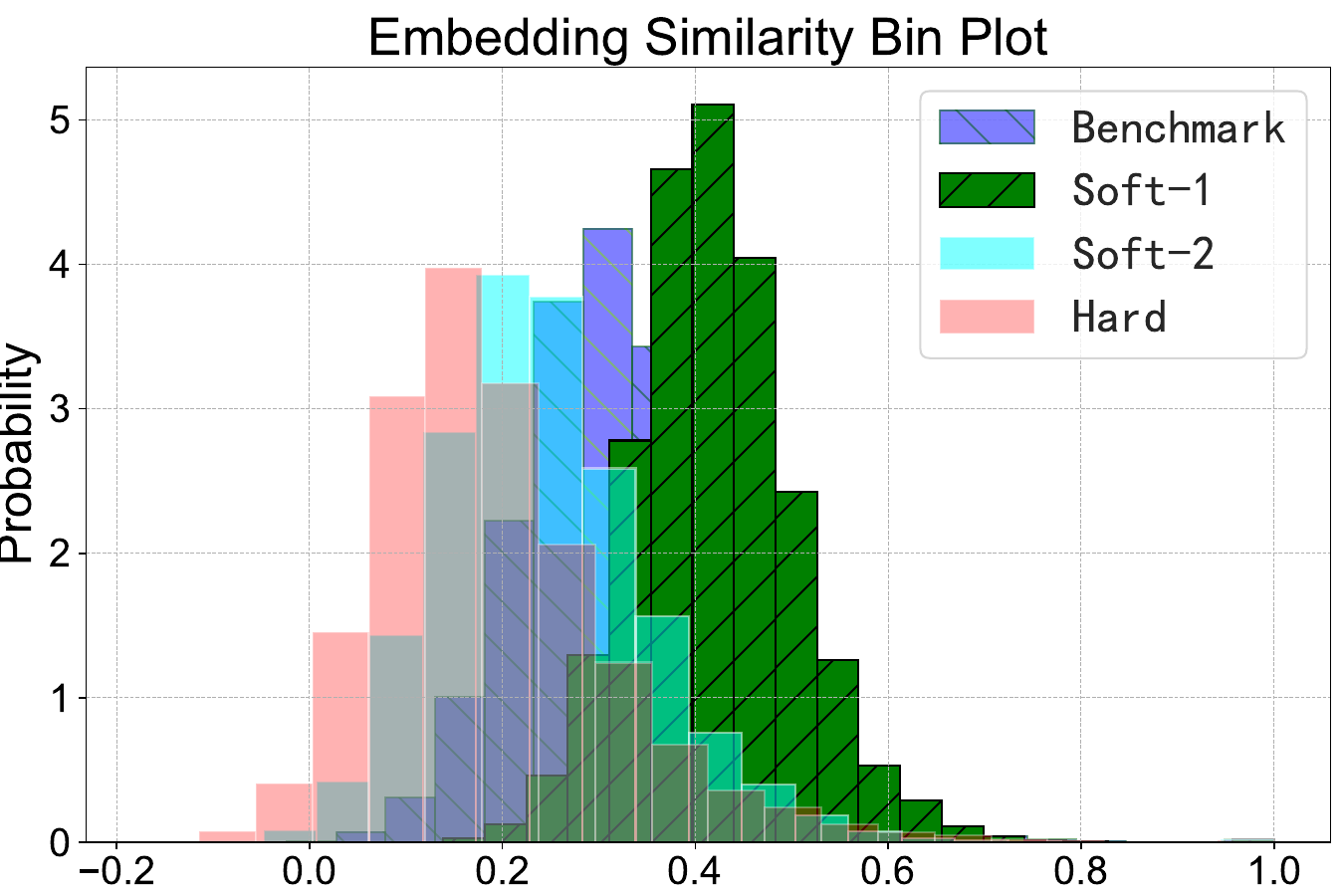}
		\captionof{figure}{Distribution of cosine similarities between the embedding vectors of different instruction texts. }
		\label{fig:distbin}
	\end{minipage}
	\hfill
	\begin{minipage}[b]{0.467\textwidth}
		\centering
		\begin{tabular}{|c|c|}
			\toprule
			Models & Distance  \\
			\midrule
			distance(Benchmark, Soft-1) & 0.6200  \\ 
			distance(Benchmark, Soft-2) & \textbf{0.2732}  \\ 
			distance(Benchmark, Hard) & 0.6882  \\ \hline 
			distance(Soft-2, Hard) & 0.1187  \\ 
			distance(Soft-1, Hard) & 2.8520  \\ \hline 
			\bottomrule
		\end{tabular}
		\captionof{table}{Symmetric KL divergences for different models. The Hard label model is the most dissimilar  from the Benchmark model, while the Soft-2 model is the closest.  This is reasonable since the Soft-2 model uses the least aggressive labels. The Soft-1 model exhibits a balance between the Soft-2 and Hard label models.}
		\label{tb:distbin}
	\end{minipage}
\end{minipage}
\end{center}
\subsection{Distributional Results}

We use LLaMA-Factory fine-tuning dataset to evaluate the distributional results.
The set contains 1,000 instructions intended for fine-tuning large language models. We generate 20,000 pairs of these instructions and evaluate the cosine similarities under different models. 
Figure~\ref{fig:distbin} shows the distribution of cosine similarities between the embedding vectors of different instruction texts. 
Table~\ref{tb:distbin} presents the symmetric KL divergences for different models. The Hard label model exhibits the greatest divergence from the Benchmark, while the Soft-2 model is the closest. 
The Soft-1 model exhibits a balance between the Soft-2 and Hard label models.
This is expected, as the Soft-2 model utilizes the least aggressive labels. 
On the other hand, there is a notable difference between the Soft-1 and Hard label models, indicating that they fine-tune towards distinct spaces.

\section{Conclusion}
In this paper, we have presented a method for improving text embeddings through contrastive fine-tuning on small datasets augmented with expert scores. Our approach leverages a limited amount of fine-tuning data, making it cost-effective and practical for real-world applications. We have demonstrated the effectiveness of our method through extensive experiments using a Q\&A dataset sourced from an online shopping website.
Our findings show that the Soft-1 and Soft-2 models, which use less aggressive labels during fine-tuning, outperform the benchmark model in terms of nDCG@10 and mAP@10 metrics, achieving higher scores and showing more stable performance across various datasets. The Hard label model, on the other hand, exhibits worse performance but a smaller standard deviation, suggesting that it may be more consistent but less effective overall.
Additionally, we have analyzed the distribution of cosine similarities between embedding vectors of different instruction texts and observed a clear distinction between "IntraSample" and "InterSample" distributions. This indicates that the embeddings are capable of effectively distinguishing between related and unrelated passages, which is crucial for retrieval tasks.
Overall, our method offers a practical solution for enhancing the quality of text embeddings in both the downstream tasks and general-purpose tasks, particularly in scenarios where labeled data is scarce. Future work could explore further improvements in the fine-tuning process and the integration of additional features/labels based on the dataset itself to enhance the representativeness and utility of the embeddings in diverse NLP applications.
Recent researches show that transformer-based models are anisotropic \citep{godey2023anisotropy, godey2024anisotropy}. It remains interesting to show whether the anisotropic issue is less severe in the high-dimensional space after fine-tuning with soft labels or not.

\setcitestyle{numbers}
\bibliography{bib}
\bibliographystyle{iclr}

\end{document}

%% file: symbols.tex
\newcommand{\gap}{\,\,\,\,\,\,\,\,}  
\mathchardef\mhyphen="2D

\newcommand\norm[1]{\left\lVert#1\right\rVert}

\def\1{\bm{1}}

\definecolor{titlepagecolor}{cmyk}{75,68,67,90}
\definecolor{titlepagecolor2}{rgb}{1.0, 0.08, 0.58}
\definecolor{emerald}{rgb}{0.31, 0.78, 0.47}
\definecolor{deeppink}{HTML}{D14064}
\definecolor{lowpink}{HTML}{ffe6ec}
 
\definecolor{lowblue}{HTML}{E1EBFE}

\let\oldforall\forall
\renewcommand{\forall}{\oldforall\, }

\usepackage{color}   

\definecolor{mylightbluetitle}{RGB}{60,113,183}
\definecolor{structurecolorblue}{RGB}{60,113,183}
\definecolor{structurecolorgreen}{RGB}{63,145,182}
\colorlet{structurecolor}{structurecolorblue}
\definecolor{structurecolorelegant}{RGB}{60,113,183}
\definecolor{structurecolorlt}{RGB}{31,119,185}

\definecolor{structurecolorHighTheoremBlue}{RGB}{220,227,248}
\definecolor{structurecolorHighTheoremGreen}{RGB}{188,222,231}
\colorlet{structurecolorHighTheorem}{structurecolorHighTheoremBlue}

\definecolor{winestain}{rgb}{0.5,0,0}
\definecolor{mydarkblue}{rgb}{0,0.08,0.45}
\definecolor{mydarkred}{rgb}{0.70,0.00,0.00}
\definecolor{mydarkgreen}{rgb}{0.00,0.30,0.00}
\definecolor{mydarkyellow}{RGB}{197,151,13}
\definecolor{mydarkpurple}{RGB}{149,18,192}
\definecolor{mydarkgray}{RGB}{64,64,64}

\definecolor{color0}  {RGB}{174,225,254} 
\definecolor{color1}  {RGB}{220,227,248} 
\definecolor{color2}  {RGB}{28,130,185} 
\definecolor{color3}  {RGB}{255,253,250} 
\definecolor{colormiddleright}  {RGB}{245,253,250} 
\definecolor{colorbottomleft}  {RGB}{255,243,250} 
\definecolor{coloruppermiddle}  {RGB}{255,253,230} 
\definecolor{colormiddleleft}  {RGB}{255,244,237}
\definecolor{colorcr}  {RGB}{249,253,232} 
\definecolor{colorreduction}  {RGB}{255,235,254} 
\definecolor{colorqr}  {RGB}{254,221,199} 
\definecolor{colorbiconjugate}  {RGB}{251,149,161} 
\definecolor{colorsvd}  {RGB}{215,247,235} 
\definecolor{colorupperright}  {RGB}{239,246,251} 
\definecolor{colorspectral}  {RGB}{206,226,243} 
\definecolor{colorbottomright}  {RGB}{220,224,236} 
\definecolor{coloreigenvalue}  {RGB}{197,203,224} 
\definecolor{colorcp} {RGB}{217, 234, 186} 
\definecolor{colorcpborder} {RGB}{233, 243, 216} 
\definecolor{colorupperleft}  {RGB}{235,243,240} 
\definecolor{colorsemidefinite}  {RGB}{217,232,226} 
\definecolor{colormiddle} {RGB}{235, 240,255}
\definecolor{colorlu}  {RGB}{220,227,255} 
\definecolor{colorals}  {RGB}{240,230,255} 
\definecolor{coloralsbkg}  {RGB}{248,243,255} 
\definecolor{canaryyellow}{rgb}{1.0, 0.75, 0.0}
\definecolor{bluepigment}{rgb}{0.0, 0.0, 1.0}
\definecolor{canarypurple}{RGB}{208, 13, 241}
\definecolor{colorGreenOcre}{RGB}{51,102,0} 
\definecolor{colorBlue2}{RGB}{200,207,248}
\definecolor{shadecolor}{gray}{0.75}

\newcommand{\bE}{\bm{E}}

\newcommand{\bp}{\bm{p}}

\newcommand{\bq}{\bm{q}}

\DeclareMathAlphabet{\mathsfit}{\encodingdefault}{\sfdefault}{m}{sl}
\SetMathAlphabet{\mathsfit}{bold}{\encodingdefault}{\sfdefault}{bx}{n}